\documentclass[11pt,letterpaper]{article}
\usepackage{naaclhlt2016}
\usepackage{times}
\usepackage{latexsym}
\usepackage{graphicx}
\usepackage[utf8]{inputenc}
\usepackage{covington}
\usepackage{amsmath}
\usepackage{url}
\usepackage{enumerate}

\naaclfinalcopy 
\def\naaclpaperid{BEA-22} 

\title{Detecting Context Dependence in
 Exercise Item Candidates Selected from Corpora}

\author{Ildikó Pilán\\
	    Swedish Language Bank, University of Gothenburg\\
	    Gothenburg, 405 30, Sweden\\
	    {\tt ildiko.pilan@svenska.gu.se}
	 }

\date{}

\begin{document}

\maketitle

\begin{abstract}
We explore the factors influencing the dependence of single sentences on their larger textual context in order to automatically identify candidate sentences for language learning exercises from corpora which are presentable in isolation. An in-depth investigation of this question has not been previously carried out. Understanding this aspect can contribute to a more efficient selection of candidate sentences which, besides reducing the time required for item writing, can also ensure a higher degree of variability and authenticity. We present a set of relevant aspects collected based on the qualitative analysis of a smaller set of context-dependent corpus example sentences. Furthermore, we implemented a rule-based algorithm using these criteria which achieved an average precision of 0.76 for the identification of different issues related to context dependence. The method has also been evaluated empirically where 80\% of the sentences in which our system did not detect context-dependent elements were also considered context-independent by human raters. 
\end{abstract}

\section{Introduction}
Extracting single sentences from corpora with the use of Natural Language Processing (NLP) tools can be useful for a number of purposes including the detection of candidate sentences for automatic exercise generation. Such sentences are also known as \textit{seed sentences} \cite{sumita2005measuring} or \textit{carrier sentences} \cite{smith2010gap} in the Intelligent Computer-Assisted Language Learning (ICALL) literature. Interest for the use of corpora in language learning has arisen already in the 1980s, since the increasing amount of digital text available enables learning through authentic language use \cite{o2007corpus}. However, since sentences in a text form a coherent discourse, it might be the case that for the interpretation of the meaning of certain expressions in a sentence, previously mentioned information, i.e. a \emph{context}, is required \cite{poesio2011computational}. Corpus sentences whose meaning is hard to interpret are less optimal to be used as exercise items \cite{kilgarriff2008gdex}, however, having access to a larger linguistic context is not possible due to copy-right issues sometimes \cite{volodina2012semi}. 

In the followings, we explore how we can automatically assess whether a sentence previously belonging to a text can also be used as a stand-alone sentence based on the linguistic information it contains. We consider a sentence \emph{context-dependent} if it is not meaningful in isolation due to: (i) the presence of expressions referring  to textual content that is external to the sentence, or (ii) the absence of one or more elements which could only be inferred from the surrounding sentences.


Understanding the main factors giving rise to context dependence can improve the trade-off between discarding (or penalizing) sub-optimal candidates and maximizing the variety of examples and thus, their authenticity. Such a system may not only facilitate teaching professionals' work, but it can also aid the NLP community in a number of ways, e.g. evaluating automatic single-sentence summaries, detecting ill-formed sentences in machine translation output or identifying dictionary examples. 

Although context dependence has been taken into consideration to some extent in previous work, we offer an in-depth investigation of this research problem. The theoretical contribution of our work is a set of criteria relevant for assessing context dependence of single sentences based on a qualitative analysis of human evaluators' comments. This is complemented with a practical contribution in the form of a rule-based system implemented using the proposed criteria which can reliably categorize corpus examples based on context dependence both when evaluated using relevant datasets and according to human raters' judgments. The current implementation of the system has been tested on Swedish data, but the criteria can be easily applied to other languages as well. 


\section{Background}
\label{sec:background}

\subsection{Corpus Examples Combined with NLP for Language Learning}
\label{bg-gdex}
In a language learning scenario, corpus example sentences can be useful both as exercise items and as vocabulary examples. Previous work on exercise item generation has adopted different strategies for carrier sentence selection. In some cases, sentences are mainly required to contain a lexical item or a linguistic pattern that constitutes the target of the exercise, but context dependence is not explicitly addressed \cite{sumita2005measuring,arregik2011automatic}. Another alternative has been using dictionary examples as carrier sentences, e.g. from WordNet \cite{pino2009semi}. Such sentences are inherently context-independent, however, they pose some limitations on the linguistic aspects to target in the exercises. In \newcite{pilan2014rule} we presented and compared two algorithms for carrier sentence selection for Swedish, using both rule-based and machine learning methods. Context dependence, which had not been specifically targeted in that phase, emerged as a key issue for sub-optimal candidate sentences during an empirical evaluation.

Identifying corpus examples for illustrating lexical items is the main purpose of the GDEX (Good Dictionary Examples) algorithm \cite{husak2010automatic,kilgarriff2008gdex} which has also inspired a Swedish algorithm for sentence selection \cite{volodina2012semi}. GDEX incorporates a number of linguistic criteria (e.g. sentence length, vocabulary frequency) based on which example candidates are ranked. Some of these are related to context dependence (e.g. incompleteness of sentences, presence of personal pronouns), but they are somewhat coarser-grained criteria not focusing on syntactic aspects. A system using GDEX for carrier sentence selection is described in \newcite{smith2010gap} who underline the importance of the well-formedness of a sentence and who determine a sufficient amount of context in terms of sentence length. \newcite{segler2007investigating} focuses on vocabulary example identification for language learners. Teachers' sentence selection criteria has been modeled with logistic regression, the main dimensions examined being syntactic complexity and similarity between the original context of a word and an example sentence.

\subsection{Linguistic Aspects Influencing Context Dependence}
\label{bg-crit}
The relationship between sentences in a text can be expressed either explicitly or implicitly, i.e. with or without specific linguistic elements requiring extra-sentential information \cite{mitkov2014anaphora}. The explicit forms include words and phrases that imply structural discourse relations or are anaphoric \cite{webber2003anaphora}. In a text, the way sentences are interconnected can convey an additional relational meaning besides the one which we can infer from the content of each sentence separately. Examples of such elements include \emph{structural connectives}: conjunctions, subjunctions and “paired” conjunctions \cite{webber2003anaphora}. 

Another form of reference to previously mentioned information is \emph{anaphora}. The phenomenon of anaphora consists of a word or phrase (\emph{anaphor}) referring back to a previously mentioned entity (\emph{antecedent}). \newcite{mitkov2014anaphora} outlines a number of different anaphora categories based on their form and location, the most common being pronominal anaphora which has also been the focus of recent research within NLP \cite{poesio2011computational,ng2010supervised,nilsson2010hybrid}. A number of resources available today have noun phrase coreference annotation, such as the dataset from the SemEval-2010 Task \cite{recasens2010semeval} and SUC-CORE for Swedish \cite{nilsson2013suc}.

Besides the anaphora categories described in \newcite{mitkov2014anaphora}, \newcite{webber2003anaphora} argue that adverbial connectives (\textit{discourse connectives}), e.g. \textit{istället} `instead', also behave anaphorically, among others because they function more similarly to anaphoric pronouns than to structural connectives. A valuable resource for developing automatic methods for handling discourse relations is the Penn Discourse Treebank \cite{prasad2008penn} containing annotations for both implicit and explicit discourse connectives. Using this resource \newcite{pitler2009using} present an approach based on syntactic features for distinguishing between discourse and non-discourse usage of explicit discourse connectives (e.g. \textit{once} as a temporal connective corresponding to "as soon as" vs. the adverb meaning "formerly"). Another phenomenon connected to context dependence is \emph{gapping} where the second mention of a linguistic element is omitted from a sentence \cite{poesio2011computational}.

\section{Datasets}
\label{sec:datasets}

Instead of creating a corpus specifically tailored for this task with gold standard labels assigned by human annotators, which can be a rather time- and resource-intensive endeavor, we explored how different types of existing data sources which contained inherently context-(in)dependent sentences could be used for our purposes.

Language learning coursebooks contain not only texts, but also single sentences in the form of exercise items, lists and language examples illustrating a lexical or a grammatical pattern. We collected sentences belonging to these two latter categories from COCTAILL \cite{volodina22you}, a corpus of coursebooks for learners of Swedish as a second language. Most exercises contained gaps which might have misled the automatic linguistic annotation, therefore they have not been included in our dataset.

Dictionaries contain example sentences illustrating the meaning and the usage of an entry. One of the characteristics of such sentences is the absence of referring expressions which would require a larger context to be understood \cite{kilgarriff2008gdex}, therefore they can be considered suitable representatives of context-independent sentences. We collected instances of good dictionary example sentences from two Swedish lexical resources: SALDO \cite{borin2013saldo} and the Swedish FrameNet (SweFN) \cite{heppin2012rocky}. These sentences were manually selected by lexicographers from a variety of corpora. 

Sentences explicitly considered dependent on a larger context are less available due to their lack of usefulness in most application scenarios. Two previous evaluations of corpus example selection for Swedish are described in \newcite{volodina2012semi} and \newcite{Pilan-Ildiko2013-9}, we will refer to these as \textsc{Eval1} and \textsc{Eval2} respectively. In the former case, evaluators including both lexicographers and language teachers had to provide a score for the appropriateness of about 1800 corpus examples on a three-point scale. In \textsc{Eval2}, about 200 corpus examples selected with two different approaches were rated by a similar group of experts based on their understandability (readability) for language learners, as well as their appropriateness as exercise items and as good dictionary examples. The data from both evaluations contained human raters' comments explicitly mentioning that certain sentences were context-dependent. We gathered these instances to create a negative sample. Since comments were optional, and context dependence was not the focus of these evaluations, the amount of sentences collected remained rather small, 92 in total. It is worth noting that this data contains spontaneously occurring mentions based on raters' intuition, rather than being labeled following a description of the phenomenon of context dependence as it would be customary in an annotation task.

The sentences from all data sources mentioned above constituted our development set. The amount of sentences per data source is presented in Table \ref{table:data}, where \textsc{CInd} indicates positive, i.e. context-independent samples, and \textsc{CDep} the negative, context-dependent ones. The suffix \textsc{-LL} stands for sentences collected from language learning materials while \textsc{-D} represents dictionary examples.

\begin{table}[ht]
\centering
\small
\begin{tabular}{|c|c|c|c|}
\hline
\bf Source & \bf Code & \bf Nr. sent & \bf Total\\
\hline
COCTAILL & \textsc{CInd-LL} & 1739 & \\
SALDO & \textsc{CInd-D} & 4305 & \bf 8729 \\
SweFN & \textsc{CInd-D} & 2685 &  \\
\hline
\textsc{Eval1} & \textsc{CDep} & 22 & \\
\textsc{Eval2} & \textsc{CDep} & 70 & \bf 92 \\
\hline
\end{tabular}
\caption{\label{table:data} Number of sentences per source.}
\end{table}
\setlength{\textfloatsep}{0.7cm}

\section{Methodology}

As the first step in developing the algorithm, we aimed at understanding the presence or absence of which linguistic elements make sentences dependent on a larger context by analyzing our negative sample. Although the number of instances in the context-independent category was considerably higher, certain linguistic characteristics of such sentences could have been connected to aspects not relevant to our task. Negative sentences on the other hand, although modest in number, were explicit examples of the target phenomenon. Information about the cultural context may also be relevant for this task, however, we only concentrated on linguistic factors which can be effectively captured with NLP tools. 

We aimed at covering a wide range of potential application scenarios, therefore we developed a method that was independent of: (i) information from surrounding sentences and (ii) the exact intended use for the selected sentences. The first choice was motivated by the fact that, even though most previous related methods (see section \ref{bg-crit}) rely on information from neighboring sentences as well, sometimes a larger context might not be available either due to the nature of the task (e.g. output of single-sentence summarization systems) or copy-right issues. Secondly, for a more generalizable approach, we aimed at assessing sentences based on whether their information content can be treated as an autonomous unit rather than according to whether they provide the appropriate amount and type of context to, for example, be solved as exercise items of a certain type. This way the method could serve as a generic basis to be tailored to specific applications which may pose additional requirements on the sentences. 

Being that the amount of negative samples was rather restricted, we opted for the qualitative method of \emph{thematic analysis} \cite{boyatzis1998transforming,braun2006using} aiming at discovering \emph{themes}, i.e.  categories, in our negative sample. Once we collected a set of context-dependent sentences, we started coding our data, in other words, manually labeling the instances with \textit{codes}, a word or a phrase shortly describing the type of element that inhibited the interpretation of the sentence in isolation (for some examples see Table \ref{table:TA_results} on the next page). In the subsequent phases, we grouped together codes into themes, i.e. broader categories, according to their thematic similarity in a mixed deductive-inductive fashion. We started out with an initial pool of themes inspired by phenomena proposed in previous literature relevant to context dependence. Some of the codes, however, could not be placed in any of these themes. For part of these we have found a theme candidate in the literature after the pattern emerged during the code grouping phase. In other cases, in absence of an existing category matching some instances of the \textsc{CDep} data, we created our own theme labels. 

\begin{table*}[ht]
\centering
\small
\begin{tabular}{|l|c|c|c|c|}
\hline
\bf Theme & \bf ID & \bf Nr & \bf Example code & \bf Example \textsc{CDep} sentence \\
\hline
Incomplete sentence&\textsc{IncompSent}&12&incorrect sent.&\it \textbf{'' p}iper hon och alla skrattar .\\
&&&tokenization&`\textbf{'' s}he whines and everyone laughs.'\\
\hline
Implicit anaphora&\textsc{ImpAnaphora}&11&omitted verb&\it Till jul skulle hon \textbf{[X]}.\\
&&&&`For Christmas she should have \textbf{[X]}.'\\
\hline
Pronominal anaphora&\textsc{PNAnaphora}&23&pronoun as&\it Eller också sitter \textbf{den} i taket.\\
&&&subject&`Or \textbf{it} sits on the roof.'\\
\hline
Adverbial anaphora 1&\textsc{AdvAnaphora1}&12&locative adverb&\it \textbf{Då} ska folk kunna lämna området .\\
(Temporal and locative) & & & & `\textbf{Then} people can leave the area.'\\
\hline
Adverbial anaphora 2&\textsc{AdvAnaphora2}&22& adv. anaphora&\it Vissa gånger sover hon inte \textbf{heller}.\\
(Discourse connectives) & & & &`Sometimes she does not sleep \textbf{either}.'\\
\hline
Structural connectives&\textsc{StructConn}&17&coordinating&\it \textbf{Men} de pratade inte på samma ställe.\\
&&& conjunction&`\textbf{But} they did not talk at the same place.'\\
\hline
Answers to closed &\textsc{CEQAnswer}&11& yes/no answer & \it \textbf{Ja,} men det är ju jul.\\
 ended questions & & & & `\textbf{Yes,} but it is of course Christmas.'\\
 \hline
Context-depend  & \textsc{CDPC} & 8 & unusual noun-  & \it Du lämnar \textbf{planen}, \textbf{tolvan}! \\
properties of concepts & & & noun comb. &`You leave the \textbf{field}, \textbf{twelve}!'\\
\hline
\end{tabular}
\caption{\label{table:TA_results} Thematic analysis results.}
\end{table*}


Besides thematic analysis, we carried out also a quantitative analysis based on the distribution of part of speech tags in both our positive and negative sample in order to identify potential differences that could support and complement the information emerged in the themes.

In the following step, we implemented a rule-based algorithm for handling context dependence using the findings from the qualitative and quantitative analyses. Since most emerged aspects could be translated into rather easily detectable linguistic clues, and a sufficiently large dataset annotated with the different context-dependent phenomena was not available for Swedish, we opted for a heuristic-based system. We applied the algorithm and observed its performance on our development data. Our primary focus was on evaluating how precisely are context-dependent elements identified in \textsc{CDep}, but we complemented this also with observing the percentage of false positives for context dependence in our positive sample. 

Finally, in order to test candidate selection empirically, a new set of sentences has been retrieved from different corpora. These sentences were then first given to our system for assessment, then the subset of candidates not containing context dependent elements were given to evaluators for an external validation.

\section{Data Analysis Results}

\subsection{Qualitative Results Based on Thematic Analysis}

The list of themes collected during our qualitative analysis is presented in Table \ref{table:TA_results}. For each theme, we provide an identifier (\emph{ID}), the number of occurrence in the \textsc{CDep} dataset (\emph{Nr\footnote{Occasionally sentences included more than one theme.}}) together with an example code and an example sentence\footnote{Tokens relevant to each theme are in bold and [X] indicates the position of an omitted element.}.

The total number of codes emerged from the data was 22, which we mapped to 8 themes. Some of the themes were related to the categories mentioned in previous literature which we described in section \ref{sec:background}. These included pronominal anaphora \cite{mitkov2014anaphora}, adverbial anaphora \cite{webber2003anaphora}, connectives \cite{miltsakaki2004annotating}. Incomplete sentences \cite{didakowski2012automatic} contained incorrectly tokenized sentences, titles and headings. Moreover, we distinguished three themes among different anaphoric expressions: pronominal anaphora, adverbial anaphora (with temporal and locative adverbs) and discourse connectives, i.e. adverbials expressing logical relations. Under the implicit anaphora theme we grouped different forms of gapping. 

Two themes that emerged from the data during the thematic analysis were answers to closed ended questions and context-dependent properties of concepts. In the case of the former category, answers were mostly of the yes/no type. As for the latter theme, our data showed that the unexpectedness of the context of a word (especially if this is short, such as a sentence) can also play a role in whether a sentence is interpretable in isolation. Previous literature \cite{barsalou1982context} defines this phenomenon as “context-dependent properties of concepts”. While the “core meanings” of words are activated “independent of contextual relevance”, context-dependent properties are “only activated by relevant contexts in which the word appears” \cite[p. 82]{barsalou1982context}. In (\ref{cdpc}) we provide an example of both context-independent and context-dependent properties of the noun \textit{tak} `roof', from the \textsc{Eval2} data.

\begin{example}
\label{cdpc}
\begin{itemize}
\item[(a)] \textit{Troligen berodde olyckan på all snö som låg på taket.}
\newline
`The accident probably depended on all the snow that covered the roof.'
\item[(b)] \textit{Fler än hundra levande kunde dras fram under taket .}
\newline
`More than a hundred [people] were pulled out from under the roof alive.'
\end{itemize}
\end{example}

Sentence (\ref{cdpc}b) was considered context-dependent by human raters, while (\ref{cdpc}a) was not. Being covered in snow (\ref{cdpc}a) appears a more easily interpretable property of roof without a larger context than having something being pulled out from under it. The context that activates the context-dependent property of roof in (\ref{cdpc}b) is that the roof had collapsed, which, however, is missing from the sentence.

Finally, for 7 sentences in our \textsc{CDep} data, no clear elements causing context dependence could have been clearly identified, these are omitted from Table \ref{table:TA_results}, but they have been preserved in the experiments. 


\subsection{Quantitative Comparison of Positive and Negative Samples}

Besides carrying out a thematic analysis, we compared our positive and negative samples also based on quantitative linguistic information in search of additional evidence for the emerged themes and to detect further aspects that could be potentially worth targeting. Overall part of speech (POS) frequency counts showed some major differences between the \textsc{CDep} and \textsc{CIndep} sentences. There was a tendency towards a nominal content in context-independent sentences, where 21.6\% of all POS tags were nouns. However, this value was 9\% lower for context-dependent sentences, which would suggest a preference for a higher density of concepts in context-independent sentences. Pronouns, on the other hand, were more frequent in context-dependent sentences (12.6\% in total) than in context-independent ones (7\% less frequent).

The qualitative analysis revealed that elements responsible for context dependence commonly occurred at the beginning of the sentence. Therefore, we compared the percentage of POS categories for this position in the two groups of sentences. Context-independent sentences showed a strong tendency towards having a noun in sentence-initial position, almost one fourth of the sentences fit into this category. On the other hand, only 3\% of the positive examples started with a conjunction, but 16\% of context-depend items belonged to this group. 

\section{An Algorithm for the Assessment of Context Dependence}
\label{sec:alg}

Inspired by the results of the thematic analysis and the quantitative comparison described above, we implemented a heuristics-based system for the automatic detection of context dependence in single sentences. 
For retrieving example sentences the system uses the concordancing API of Korp \cite{borin2012korp}, a corpus-query system giving access to a large amount of Swedish corpora. All corpora were annotated for different linguistic aspects including POS tags and dependency relation tags which served as a basis for the implementation. The system scores each sentence based on the amount of phenomena detected that match an implemented context dependence theme.
Users can decide whether to \emph{filter}, i.e. discard sentences that contain any element indicating context dependence. Alternatively, sentences can be \emph{ranked} according to the amount of context-dependent issues detected: sentences without any such elements are ranked highest, followed by instances minimizing these aspects. 
All themes have an equal weight of 1 when computing the final ranking score, except for pronominal anaphora in which case, if pronouns have antecedent candidates, the weight is reduced to 0.5. In the followings, we provide a detailed description of the implementation of the themes listed in Table \ref{table:TA_results}.
\newline

\noindent
\textbf{Incomplete sentence.}
To detect incomplete sentences the algorithm scans instances for the presence of an identified dependency root, the absence of which is considered to cause context dependence. Moreover, orthographic clues denoting sentence beginning and end are inspected. Sentence beginnings are checked for the presence of a capital letter optionally preceded by a parenthesis, quotation mark or a dash, frequent in dialogues. Sentences beginning with a digit are also permitted. Sentence end is checked for the presence of major sentence delimiters (e.g. period, exclamation mark). 
\newline


\noindent
\textbf{Implicit anaphora.}
Candidate sentences are checked for gapping, in other words, omitted elements. Our system categorizes as gapped (elliptic) a sentence which either lacks a finite verb or a subject. Finite verbs are all verbs that are not infinite, supine or participle. Modal verbs are considered finite in case they form a verb group with another verb. Subjects include also logical subjects, and in the case of a verb in imperative mode, no subject is required.
\newline
\newline

\noindent
\textbf{Explicit pronominal anaphora.}
We considered in this category the third person singular pronouns \textit{den} `it' (common gender) and \textit{det} `it' (neuter gender) as well as demonstrative pronouns (e.g. \textit{denna} `this', \textit{sådan} `such' etc.). We did not include here the animate third person pronouns \textit{han} `he' and \textit{hon} `she' since corpus-based evidence suggests that these are often used in isolated sentences in coursebooks \cite{scherrer2007rivstart} as well as in conversation \cite{mitkov2014anaphora}.
Similarly to the English pronoun \textit{it}, the Swedish equivalent \textit{det} can also be used non-anaphorically in expositions, clefts and expressions describing a local situation, such as time and weather \cite{holmes2003swedish,li2009identification,gundel2005pronouns} as the examples in (\ref{det}) show.

\begin{example}
\label{det}
\begin{itemize}
\item[(a)] \textit{det} with weather-related verbs
\newline
\textit{Det regnar.}
\newline
`It is raining.'
\item[(b)] Cleft
\newline
\textit{Det är sommaren (som) jag älskar.}
\newline
`It is the summer (that) I like.'
\item[(c)] Exposition
\newline
\textit{Det är viktigt att du kommer.}
\newline
`It is important that you come.'
\end{itemize}
\end{example}

Our system treats as non-anaphoric the pronoun \textit{det} if it is expletive (pleonastic) syntactically according to the output of the dependency parser which covers expositions and clefts. To handle cases like (2a), weather-related verbs have been collected from lexical resources. The list currently comprises 14 items. First, verbs related to the class \emph{Weather} in the Simple+ lexicon \cite{kokkinakis2000annotating} have been collected. Then for each of these, the child nodes from the SALDO lexicon have been added. Finally, the list has been complemented with a few manual additions. 

For potentially anaphoric pronouns, the system tries to identify antecedent candidates in a similar way to the robust pronoun resolution algorithm proposed in \newcite{mitkov1998robust}. We count proper names and nouns occurring with the same gender and number to the left of the anaphora. This is complemented with an infinitive marker headed by a verb as potential candidate for \textit{det}. Since certain types of information useful for antecedent disambiguation were not available through our annotation pipeline or lexical resources for Swedish (e.g. gender for named entities, animacy), the final step for scoring and choosing candidates is not applied in this initial version of the algorithm. Lastly, pronouns followed by a relative clause introduced by \textit{som} `which' were considered non-anaphoric.  
\newline

\noindent
\textbf{Explicit adverbial anaphora.}
Adverbs emerged as an undesirable category during both \textsc{Eval1} and \textsc{Eval2}. However, a deeper analysis of our development data revealed that not all adverbs have equal weight when determining the suitability of a sentence. Some are more anaphoric then others. We collected a list of anaphoric adverbs based on \newcite{teleman1999svenska}. Certain time and place adverbials, also referred to as demonstrative pronominal adverbs \cite{webber2003anaphora} are used anaphorically (e.g. \textit{där} `there', \textit{då} `then'). Sentences containing these adverbs are considered context-independent only when: (i) they are the head of an adverbial of the same type that further specifies them, e.g. \textit{där på landet} `there on the countryside'; (ii) they appear with a determiner, which in Swedish builds up a demonstrative pronoun, e.g. \textit{det där huset} `that house'.
\newline

\noindent
\textbf{Discourse connectives.}
Discourse connectives, i.e. adverbs expressing logical relations, fall usually into the syntactic category of conjunctional adverbials in the dependency parser output. Several conjunctional adverbials appear in the context-dependent sentences from \textsc{Eval1} and \textsc{Eval2}. Our system categorizes a sentence containing a conjuctional adverb context-independent when a sentence contains:
(i) at least 2 coordinate clauses; (ii) coordination or subordination at the same dependency depth or a level higher, that is, a sibling node that is either a conjunction or a subjunction.
\newline

\noindent
\textbf{Structural connectives.}
Sentences with conjunctions as dependency roots are considered context-dependent unless they are paired conjunctions with both elements included (e.g. \textit{antingen ... eller} `either ... or'). Conjunctions in sentence initial position are also treated as an indication of context dependence except when there are at least two clauses or conjuncts in the sentence.

\noindent
\textbf{Answers to closed ended questions.}
To identify sentences that are answers to closed ended questions, the algorithm tries to match POS-tag patterns of sentence-initial interjections (e.g. \textit{ja} `yes', \textit{nej} `no') and adverbs surrounded by minor delimiters (e.g. dash), the initial delimiter being optional in the case of interjections. 
\newline

\noindent
\textbf{Context-dependent properties of concepts.}
Apart from the theme implementations described above, we are currently investigating the usefulness of word co-occurrence information for this theme. The corpus query tool Korp for instance offers an API providing mutual information scores. The intuition behind this idea is that the frequency of words appearing together is positively correlated with the unexpectedness of the association between them.
\newline


\section{Performance on the Datasets}

We evaluated our system both on the hand-coded negative example sentences collected from \textsc{Eval1} and \textsc{Eval2} (\textsc{CDep}) and the positive samples comprised of the good dictionary examples (\textsc{CIndep-D}) and the coursebook sentences (\textsc{CIndep-LL}). The performance when predicting different aspects of context dependence is presented in Table \ref{table:precision}. 

\begin{table}[ht]
\centering
\small
\begin{tabular}{|c|ccc|}
\hline
\bf Theme & \bf Precision & \bf Recall & \bf F1 \\
\hline
\textsc{IncompSent} &0.75 &0.5 &0.6 \\
\textsc{ImpAnaphora}& 0.33 &0.36 &0.35\\
\textsc{PNAnaphora} & 0.75 &0.78 &0.77\\
\textsc{AdvAnaphora1}& 0.91 &0.83&0.87\\
\textsc{AdvAnaphora2}&0.87 &0.59&0.70\\
\textsc{StructConn} &0.7 &0.82 &0.76\\
\textsc{CEQAnswer} & 1.0 &0.55&0.71\\
\hline
Average & \bf 0.76& 0.63 &0.60\\
\hline
\end{tabular}
\caption{\label{table:precision} Theme prediction performance in \textsc{CDep} sentences.}
\end{table}

We focused on maximizing precision, i.e. on correctly identifying as many themes as possible in the hand-coded \textsc{CDep} sentences, recall values were of lower importance since we aimed at avoiding every context-dependent sentence rather than retrieving them all. Most themes were correctly identified, all themes except one was predicted with a precision of at least 0.7 and above. The only theme that yielded a lower result was that of implicit anaphoras. The error analysis revealed that these cases were mostly connected to an incorrect dependency parse of the sentences, mainly subjects tagged as objects in sentences with an inverted (predicate-subject) word order.

As mentioned previously, we strived for minimizing sub-optimal sentences in terms of context dependence, while trying to avoid being excessively selective to maintain a varied set of examples. To assess performance with respect to this latter aspect, we inspected also the percentage of sentences identified as context-dependent in dictionary examples (\textsc{CInd-D}) and coursebook sentences (\textsc{CInd-LL}). The percentage of predicted themes per dataset is shown in Table \ref{table:cl_results} where \emph{Total} stands for the percentage of sentences with at least one predicted theme.

\begin{table}[ht]
\centering
\small
\begin{tabular}{|c|c|c|}
\hline
\bf Theme & \bf \textsc{CInd-D} & \bf \textsc{CInd-LL} \\
\hline
IncompSent  & 2.37 & 3.39\\
ImpAnaphora  & 4.61 & 5.80\\
PNAnaphora  & 9.39 & 11.0\\
AdvAnaphora1 & 3.59 & 2.93\\
AdvAnaphora2  & 9.95 & 3.74\\
StructConn & 3.70 & 0.92\\
CEQAnswer & 0.37 & 2.59\\
\hline
\bf Total & \bf 33.35 & \bf 26.74\\
\hline
\end{tabular}
\caption{\label{table:cl_results} Percentage of sentences with a predicted theme in the \textsc{CInd} datasets.}
\end{table}

We can observe that even though all sentences are expected to be context-independent, our system labeled as context-dependent about three out of ten good dictionary examples and coursebook sentences. The error analysis revealed that some of these instances did indeed contain context-dependent elements, e.g. the conjunction \textit{men} `but' in sentence-initial position. In \textsc{CInd-LL} in the case of some sentences containing anaphoric pronouns an image provided the missing context in the coursebook, thus not all predicted cases were actual false positives, but rather, they indicated  some noise in the data. As for dictionary examples, the presence of such sentences may also suggest that the criterion of context dependence can vary somewhat depending on the type of lexicon or lexicographers' individual decisions.

Some sentences exhibited more than one phenomenon connected to context dependence. Multiple themes were predicted in 30.43\% of the \textsc{CDep} sentences, but only 6.54\% and 7.25 of the \textsc{CInd-D} and \textsc{CInd-LL} sentences respectively.

\section{User-based Evaluation Results}

The algorithm was tested also empirically during an evaluation of automatic candidate sentence selection for the purposes of learning Swedish as a second language. The evaluation data consisted of 338\footnote{We excluded 8 sentences with incomplete evaluator scores during the calculation of the results.} sentences retrieved from a variety of modern Swedish corpora and classified as not containing context dependence themes according to our algorithm (with the exception of 4 control sentences that were context-dependent). These were all unseen sentences not present in the datasets described in section \ref{sec:datasets}. In the evaluation setup, all implemented themes were used as filters, i.e. sentences containing any recognized element connected to context dependence, described in section \ref{sec:alg}, were discarded. Besides context dependence, the evaluated system incorporated also other selection criteria (e.g. readability), but for reasons of relevance and space these aspects and the associated results are not discussed here.

The selected sentences were given for evaluation to 5 language teachers who assessed the suitability of these sentences based on 3 criteria: (i) their degree of being independent of context, (ii) their CEFR\footnote{The Common European Framework of Reference for Languages  (CEFR) is a scale describing proficiency levels for second language learning \cite{councilofeurope2001}.} level and (iii) their overall suitability for language learners. Teachers were required to assess this latter aspect without a specific exercise type in mind, but considering a learner reading the sentence instead. Sentences were divided into two subsets, each being rated by at least 2 evaluators. Teachers had to assign a score between 1 to 4 to each sentence according to the scale definition in Table \ref{table:eval_scale}.

\begin{table}[ht]
\centering
\begin{tabular}{|c|l|}
\hline
\multicolumn{2}{|l|}{\it The sentence...}\\
\hline
1 & \it ... doesn't satisfy the criterion. \\
2 & \it ... satisfies the criterion to a smaller extent.\\
3 & \it ... satisfies the criterion to a larger extent.\\
4 & \it ... satisfies the criterion entirely.\\
\hline
\end{tabular}
\caption{\label{table:eval_scale} Evaluation scale. }
\end{table}


The results were promising, the average score over all evaluators and sentences for context independence was 3.05, and for overall suitability 3.23. For context-independence, 61\% of the sentences received score 3 or 4 (completely satisfying the criterion) from at least half of the evaluators, and 80\% of the sentences received an average score higher than 2.5. This latter improves significantly on the percentage of context-dependent sentences that we reported previously in \newcite{Pilan-Ildiko2013-9}, where about 36\% of all selected sentences were explicitly considered context-dependent by evaluators. 

Furthermore, we computed the Spearman correlation coefficient for teachers' scores of overall suitability and context dependence to gain insight into how strongly associated these two aspects were according to our evaluation data. The correlation over all sentences was $\rho$=0.53, which indicates that not being context-dependent is positively associated with overall suitability. Therefore, context dependence is worth targeting when selecting carrier sentences. 

\section{Conclusion and Future Work}
We described a number of criteria that influence context dependence in corpus examples when presented in isolation. Based on the thematic analysis of a set of context-dependent sentences, we implemented a rule-based algorithm for the automatic assessment of this aspect which has been evaluated not only on our datasets but also with the help of language teachers with very positive results. 

About 76\% of themes were correctly identified in context-dependent sentences, while the amount of false positives in the context-independent data was maintained rather low. 
Approximately 80\% of candidate sentences selected with a system incorporating the presented algorithm were deemed context-independent in our user-based evaluation. The results also showed a positive correlation between sentences being context-independent and overall suitable for language learners. 

In the future, we are planning to explore the extension of the algorithm to other languages as well as to experiment with machine learning approaches for this task using, among others, the resources mentioned in this paper. 

\bibliography{naaclhlt2016}
\bibliographystyle{naaclhlt2016}

\end{document}